\documentclass[runningheads]{llncs}

\usepackage{eccv}

\usepackage{eccvabbrv}

\usepackage{graphicx}
\usepackage{booktabs}
\usepackage{pifont}
\usepackage{mathrsfs}
\usepackage{maths}
\usepackage{arydshln}
\usepackage{multirow}
\usepackage{threeparttable}
\usepackage{amsmath}        
\usepackage{algorithm}      
\usepackage{algpseudocode}  
\usepackage{siunitx}
\usepackage{wrapfig}
\usepackage{makecell}

\usepackage[accsupp]{axessibility}  

\def\method{Fed-Protoid} 
\def\setting{DUDA-Rid}

\usepackage{hyperref}

\usepackage{orcidlink}

\begin{document}

\title{Privacy-Preserving Adaptive Re-Identification without Image Transfer} 

\titlerunning{Privacy-Preserving Adaptive Re-Identification without Image Transfer}

\author{Hamza Rami\inst{1,2}\orcidlink{0009-0002-3654-3878} \and
Jhony H. Giraldo\inst{1}\orcidlink{0000-0002-0039-1270} \and
Nicolas Winckler\inst{2}\and
Stéphane Lathuilière\inst{1}}

\authorrunning{H. Rami et al.}

\institute{LTCI, Télécom Paris, Institut Polytechnique de Paris \and Atos}

\maketitle
\begin{abstract}
  Re-Identification systems (Re-ID) are crucial for public safety but face the challenge of having to adapt to environments that differ from their training distribution.
   Furthermore, rigorous privacy protocols in public places are being enforced as apprehensions regarding individual freedom rise, adding layers of complexity to the deployment of accurate Re-ID systems in new environments.
    For example, in the European Union, the principles of ``\textit{Data Minimization}'' and ``\textit{Purpose Limitation}'' restrict the retention and processing of images to what is strictly necessary. 
    These regulations pose a challenge to the conventional Re-ID training schemes that rely on centralizing data on servers.
    In this work, we present a novel setting for privacy-preserving Distributed Unsupervised Domain Adaptation for person Re-ID (\setting) to address the problem of domain shift without requiring any image transfer outside the camera devices.
   To address this setting, we introduce \method, a novel solution that adapts person Re-ID models directly within the edge devices.
    Our proposed solution employs prototypes derived from the source domain to align feature statistics within edge devices.
    Those source prototypes are distributed across the edge devices to minimize a distributed Maximum Mean Discrepancy (MMD) loss tailored for the \setting~setting.
    Our experiments provide compelling evidence that \method~outperforms all evaluated methods in terms of both accuracy and communication efficiency, all while maintaining data privacy.
    \keywords{Person Re-ID \and Unsupervised Domain Adaptation \and Federated Learning}
\end{abstract}    
\section{Introduction}
\label{sec:intro}

Person Re-Identification (\textit{Re-ID}) is a crucial task in computer vision, aimed at identifying specific individuals from a collection of images taken by various cameras \cite{Ye2022DeepLF}. 
The ability to perform Re-ID accurately and efficiently is essential for advancing intelligent surveillance systems and enhancing public safety.
Recent years have witnessed remarkable progress in Re-ID performance, thanks to the adoption of deep learning techniques \cite{CCS}. 
However, applying these approaches to data that is visually different from their training set results in a performance drop \cite{BoT}.
Annotating new data for each distinct environment is often infeasible, prompting previous studies to introduce Unsupervised Domain Adaptation (\textit{UDA}) methods for person Re-ID.

\begin{figure}[tb]
    \centering
    \begin{subfigure}[t]{0.5\textwidth}
        \centering
        \includegraphics[height=0.9in]{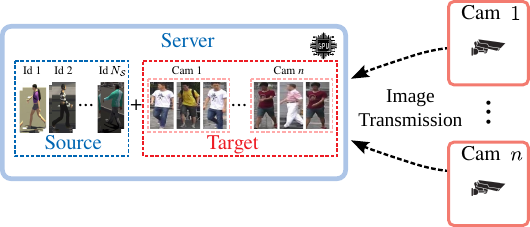}
        \caption{Traditional UDA for Person Re-ID (UDA-Rid)\label{fig:uda}}
    \end{subfigure}%
    ~ 
    \begin{subfigure}[t]{0.5\textwidth}
        \centering
        \includegraphics[height=1.in]{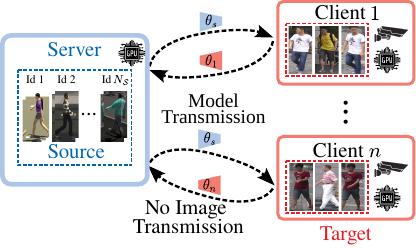}
        \caption{Privacy-preserving Distributed UDA for person Re-ID (DUDA-Rid)\label{fig:teaserDUDA}}
    \end{subfigure}

   \caption{In traditional Unsupervised Domain Adaptation (UDA) as depicted in Fig. (a), images are transmitted to a centralized server, which combines the unlabeled target images with the annotated source samples to train a model. In contrast, Distributed UDA for person re-identification (DUDA-Rid) shown in Fig. (b) keeps target images exclusively on edge devices. The learning process is divided between the server and cameras, the latter being equipped with local computational resources (\raisebox{-.3ex}{\includegraphics[height=0.3cm]{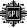}}). 
   Only model parameters are exchanged between the clients and the server.
   }
    \label{fig:teaser}
\end{figure}

UDA methods \cite{MMT, SpCL, PTGAN}, combine a well-annotated dataset (\textit{source domain}) with an unlabeled dataset (\textit{target domain}), as illustrated in Fig.~\ref{fig:uda}.
The objective of UDA methods is to train a model that can perform effectively in a new environment.
Despite remarkable advancements in recent years~\cite{MMT,SpCL}, applying UDA to person Re-ID (\textit{UDA-Rid}) encounters privacy concerns due to the need to collect and store images of individuals in public areas.
Rigorous privacy regulations in many countries restrict technology providers from retaining images of people. 
For example, within the European Union, the General Data Protection Regulation (GDPR) obligates technology providers to adhere to the principles of ``\textit{Data Minimization}'' \cite{GDPR_Ch2_Art5c} and ``\textit{Purpose Limitation}'' \cite{GDPR_Ch2_Art5b}, requiring that personal data be processed only when it is necessary for a designated purpose.
These general principles prompt the following question: \textit{What minimal data usage is truly ``necessary'' for Re-ID systems?}


An initial answer to this question emerges from recent studies~\cite{cvprw,Rami_2024_WACV} that have developed approaches for UDA-Rid, focusing on eliminating the necessity for storing images. These approaches align with privacy regulations thereby clarifying GDPR’s practical implications. 
However, these methods typically require transferring all captured images to a central server, which also poses privacy challenges \cite{GDPR_Ch5_Art44_49}.
Our work explores an alternative perspective on the question of minimal data usage: \textit{Is transferring images outside the cameras truly ``necessary'' for Re-ID?}
Our goal is to demonstrate that adaptation can be performed exclusively within edge devices, ensuring no image data is transmitted beyond its capture point as illustrated in Fig. \ref{fig:teaserDUDA}.
This paradigm provides a privacy-compliant solution while leveraging the benefits of advanced Re-ID models. 

To avoid the need for transmitting images, we approach this privacy-preserving Distributed UDA for person Re-ID (\setting) task as a federated learning problem which inherently entails two interconnected challenges:
(i) training the model in a distributed setup, and
(ii) addressing the domain gap between the source and target datasets.
Therefore, the key challenge behind the proposed setting is to simultaneously tackle the domain gap while working within a federated learning framework.

To jointly address the privacy and domain shift challenges in \setting, we introduce a novel Federated Prototype-based learning for person Re-ID (\textit{\method}) algorithm that enables domain adaptation without transmitting any image over the camera network.
\method~integrates a pseudo-labeling framework within the federated learning setup, and we propose a distributed version of the Maximum Mean Discrepancy (\textit{MMD}) technique to enhance alignment between the source and target domains.
Usually, MMD is calculated in a reproducing kernel Hilbert space using the kernel trick, which involves comparing source and target samples.
Instead, we compute source prototypes and only share these prototypes with clients to adhere to privacy constraints.
This approach for domain adaptation achieves high adaptation capabilities while keeping communication requirements to a minimum. 
\method~readily outperforms all evaluated methods for \setting~in various challenging conditions in real-to-real and synthetic-to-real tasks.
Furthermore, we show that using self-supervised pre-training \cite{LUP} coupled with a Vision Transformer (\textit{ViT}) significantly enhances performance across most scenarios for \setting. 
We refer to this architecture as \method++.

Our main contributions can be summarized as follows:
\begin{itemize}
    \item To our knowledge, we are the first to introduce and address the \setting~problem.
    \item We introduce a novel \method~algorithm that uses prototypes to jointly address distributed learning and domain shift in \setting.
    To this end, we propose a distributed version of the MMD loss to solve the domain gap in the federated setting.
    \item We further propose a \method++, which uses ViT and recent self-supervised pre-training techniques to achieve additional gains\footnote{Code available:  \href{https://github.com/ramiMMhamza/Fed-Protoid}{https://github.com/ramiMMhamza/Fed-Protoid}}.
\end{itemize}
\section{Related Work}
\label{sec:related}


\noindent\textbf{Domain adaptation for person Re-ID.} The current methods for domain adaptation can be broadly classified into three categories.
The first is the \emph{domain translation-based} methods~\cite{ATN,SDA,IGC}, which use style transfer techniques such as CycleGAN~\cite{CycleGan} to modify the source domain to match the appearance of the target set. 
Recent studies in this category have focused on enhancing the translation process via self-similarity preservation~\cite{SPGAN} or camera-specific translation~\cite{HHL}.
These types of methods are not well-suited for the \setting~problem since current federated learning methods with generative models are limited to toy datasets such as MNIST or CIFAR-10 \cite{rasouli2020fedgan,kortocci2022federated}.


The second category is based on \textit{domain-invariant} feature learning. Shan \etal \cite{lin2018multi} proposed a framework for Re-ID by minimizing the distribution variation of the source's and target's mid-level features based on the MMD loss. Huang \etal \cite{huang2019domain} designed a novel domain adaptive module to separate the feature map, while Liu \etal \cite{liu2020domain} introduced a coupling optimization method for domain adaptive person Re-ID. 
Despite their effectiveness, these methods assume unrestricted access to the target domain on the server, relying on continuous image transmission and storage between cameras and the central server, an assumption that conflicts with privacy constraints in real-world applications.

The third category is the \emph{pseudo-labeling} methods that utilize an iterative process alternating between clustering and fine-tuning \cite{theory&practice, DGM, BottomUp, CPS, delorme2021canu}. 
Fan \etal \cite{SB} finetuned the Re-ID model using the cluster indexes as labels.
Several extensions have been made to this framework such as Self-Similarity Grouping (\textit{SSG}) \cite{SSG}, Mutual-Mean Teaching (\textit{MMT}) \cite{MMT}, and Self-paced Contrastive Learning (\textit{SpCL}) \cite{SpCL}.
SSG \cite{SSG} assigns different pseudo-labels to global and local features, MMT \cite{MMT} employs a teacher-student framework with two student networks, and SpCL \cite{SpCL} gradually constructs more reliable clusters to refine a hybrid memory containing both source and target images.
We opt for the pseudo-labeling framework as it outperforms previous techniques on most datasets and since it is compatible with our \setting~setting.
Nevertheless, naively using a pseudo-labeling framework like MMT in the federated scenario incurs high communication costs.
Therefore, we design our approach to reduce communication requirements between the clients and the server.
Furthermore, our pseudo-labeling approach is enhanced with an explicit feature alignment mechanism based on MMD minimization.



\noindent\textbf{Federated learning for person Re-ID.} Federated Learning (FL)~\cite{mcmahan} aims at learning separately from multiple models trained on edges local data. 
FL restricts the sharing of data between clients and the server, as well as between clients to protect data privacy.
FL has been applied to various computer vision tasks like image segmentation \cite{FedDG}, classification \cite{FedProx}, and person Re-ID \cite{FedPav}. 
Federated Averaging (\textit{FedAvg}) \cite{mcmahan} was first proposed by McMahan \etal based on averaging local models trained with local data and redistributing the averaged server model to the edges. 
Since FedAvg requires all the models in the edges to be identical to the server model, Federated Partial Averaging (\textit{FedPav}) \cite{FedPav} was proposed to leverage only the common part of the clients' models (for example the backbones).
In this work, we adapt the FedPav to include also the weights of the model being trained on the labeled source domain.

FL has also been investigated in the task of person Re-ID.
FedReID \cite{FedReID} was first proposed to solve the task of supervised person Re-ID, which incorporates the FedPav optimization technique.
A second work that also tackles the problem of FL in person Re-ID is FedUnReID \cite{FedUnReID}, where the authors proposed an adaptation of the well-known unsupervised baseline for person Re-ID BUC \cite{BottomUp}. 
In this spirit, FedUCA \cite{FedUca} was recently introduced to address the challenge of FL for person Re-ID.
The authors draw inspiration from CAP \cite{CAP}, adopting both inter- and intra-camera losses to update a memory bank for each client.
These methods focus on the setting of federated by dataset. 
This setting represents client-edge architecture, where clients are defined as the edge servers.
Each edge server collects and processes images from a network of multiple cameras.
In contrast, our work focuses on a more restricted federated setting which does not allow the transmission of images between the cameras and any edge server. 
Finally, adapting FedUCA to our context is impractical.
This is because, in our setting, each client possesses images from a single camera device, rendering the optimization of the inter-camera loss unfeasible.


\noindent\textbf{Prototypical learning.}
The concept of prototypes in modern machine learning was first introduced in the field of few-shot learning to learn a metric space where classification can be performed by computing distances to prototype representations of each class \cite{snell2017prototypical}. 
Following this spirit, prototypical networks were applied to various computer vision tasks, such as semantic segmentation \cite{dong2018few,nguyen2019feature} and continual learning \cite{rebuffi2017icarl,han2020continual}.
Prototypical learning has also made its way into federated learning, initially applied to diverse domains unrelated to person Re-ID.
For instance, Federated Prototype learning (\textit{FedProto}) \cite{tan2022fedproto} strives to align features globally using prototypes.
Classifier Calibration with Virtual Representations (\textit{CCVR}) \cite{luo2021no} generates virtual features by leveraging an approximated Gaussian mixture model. 
More recently, Federated Prototypes Learning (\textit{FPL}) \cite{huang2023rethinking} incorporates cluster prototypes and unbiased prototypes to mitigate the domain gap between the data in the server and clients.
Notably, these previous methods are tailored for scenarios where prior information about the number of classes is available, such as in MNIST and CIFAR-10 datasets.
\method~is the first attempt to leverage prototypes in \setting, which brings new challenges due to the unsupervised nature of the problem.


\section{Federated Prototype-based Re-ID}
\label{sec:method}

\noindent \textbf{Problem definition}. 
The objective of this work is to train a model $F_\param$ with parameters $\param$ to identify individuals in a collection of $n$ cameras deployed in a target environment.
To this end, we have at our disposal $n$ unlabeled datasets $\{\mathcal{D}_1,\mathcal{D}_2,\dots,\mathcal{D}_n\}$ associated to each camera-client. 
Each dataset is composed of $N_i$ training samples (images): $\mathcal{D}_i = \{\im^{(i)}_j \}_{j=1}^{N_i}$. 
Each target dataset $\mathcal{D}_i$ is confined to its respective edge camera device and cannot be transmitted, with each camera functioning as a client that interacts solely with a centralized server.
We also have an annotated source dataset \(\mathcal{S}=\{(\im^{\mathcal{S}}_j, \vy^{\mathcal{S}}_j)\}_{j=1}^{N_s}\) available on the server, where \(N_s\) represents the number of instances in the source dataset.
The main challenge in this \setting~setting is to align the distributions of the different clients with the source domain in a distributed and privacy-preserving manner, \ie, without sharing images at any point.


In classical UDA-Rid joint learning, the training objective commonly involves two main loss terms: the source domain loss $\mathcal{L}_s$, and the target domain loss $\mathcal{L}_t$.
In non-distributed UDA-Rid, learning is commonly performed via the minimization of a linear combination of both source and target domain datasets as follows:
\begin{equation}
    \mathcal{L}(\param) =  \mathbb{E}_{(\vx,\vy) \sim \mathcal{S}}\mathcal{L}_s(\param,\vx,\vy) + \mathbb{E}_{\vx \sim \mathcal{D}}\mathcal{L}_t(\param,\vx),
\end{equation}
where $\mathcal{D} = \bigcup_{i=1}^n \mathcal{D}_{i}$.
Typically, this loss is minimized using stochastic gradient descent.
However, in our \setting~setting, the gradient of this total loss cannot be estimated without important communication costs. 
This is because the source term can be accessible only on the server via the source model, which we designate as the \textit{pseudo-client}.
Meanwhile, each device \(i\) is limited to compute only its local target loss term: \(\mathbb{E}_{\vx \sim \mathcal{D}_i}\mathcal{L}_t(\param,\vx)\).
In the following, we outline our training strategy to minimize the total loss $\mathcal{L}$ in a distributed manner.
Additionally, we describe the specifications of each loss term to facilitate communication-efficient and robust learning.


\subsection{Overview of \method}

\begin{figure}[tb]
    \centering
    \includegraphics[width=\textwidth]{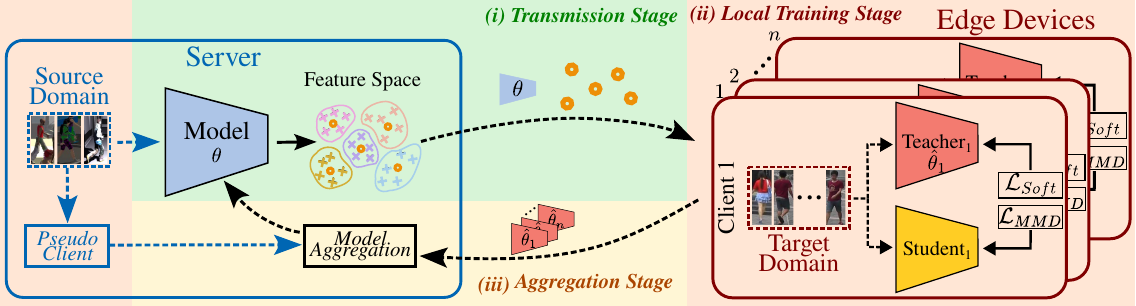}
    \caption{\textbf{The pipeline of \method.} Our algorithm aggregates $n$ edge-client models and one pseudo-client model in the server. Source prototypes are computed with the aggregated model.
    The prototypes and aggregated model are then distributed to all edge devices for local unsupervised training.
    This local training on each client involves cross-entropy, triplet, and Maximum Mean Discrepancy (MMD) loss functions.}
    \label{fig:pipeline}
\end{figure}

Figure \(\ref{fig:pipeline}\) shows the pipeline of \method~for the \setting~setting. 
Our algorithm aggregates \(n\) client models along with the pseudo-client in a distributed setting.
It adheres to standard practices in FL and functions in rounds. 
Each round is composed of three stages: (i) \textbf{transmission stage}: the aggregated model is distributed to every client and pseudo-client; (ii) \textbf{local training stage}: each client, as well as the pseudo-client, adapts their local model; (iii) \textbf{aggregation stage}: the local models are transmitted back to the server for aggregation.

At the beginning of each new round, the \textit{transmission stage} also includes the transfer of source prototypes that are later used for source-target alignment.
The aggregated model $F_\param$ computes the features of the source samples and the prototypes of each individual as the centroid of its feature representations.
The prototypes of $\mathcal{S}$ are then transmitted along with the aggregated model $F_\param$ to all clients. 
Note that we assume the server utilizes either synthetic data or real data gathered in compliance with relevant legislation. 
Consequently, the transmission of source prototypes does not breach the privacy-preserving constraints.

In the \textit{local training stage}, we use a teacher-student architecture to adapt $\param$ to the unlabeled target dataset $\mathcal{D}_i$ on each device $i$, and to the labeled source dataset $\mathcal{S}$.
The server updates the pseudo-client via supervised training, while the local adaptation on each client involves cross-entropy, triplet, and MMD loss functions.
Considering the use of the cross-entropy loss and the variation of the number of identities for each client, we add to each local device $i$ a personalized classifier head $\mathcal{C}_i$. 
This classifier is designed to match the number of classes to the respective number of identities in each client, including the pseudo-client.


Finally, in the \textit{aggregation stage}, the server gathers and aggregates the $n$ client models $\{ F_{\hat{\param}_1}, F_{\hat{\param}_2}, \dots, F_{\hat{\param}_n} \}$ obtained in the \emph{local training stage} and the model $F_{\hat{\param}_s}$ trained on the source dataset using a weighted average sum as follows:
\begin{equation}
    \param = \alpha \hat{\param}_s + (1 - \alpha)  \sum_{i=1}^{n} w_i \hat{\param}_i,
    \label{aggregation_rule}
\end{equation}
where $\{ \hat{\param}_s, {\hat{\param}_1}, {\hat{\param}_2}, \dots, {\hat{\param}_n} \}$ are the parameters of the client models after adaptation, $\alpha$ is the weight contribution of the pseudo-client model $\hat{\param}_s$, and $w_i$ is the weight assigned to the $i$th client model given by $w_i= \frac{N_i}{\sum_{i=1}^{n} N_i}$.

\subsection{Teacher-student architecture}
\label{sec:teacher-student}

All clients, encompassing the pseudo-client, employ the same teacher-student architecture. 
This framework is chosen for its effectiveness in enabling self-training techniques, which have been shown to yield optimal performance in UDA-Rid scenarios. 
While self-training is not required in the source domain due to its labeled nature, the use of the teacher-student framework favors similar training dynamics across both clients and the pseudo-client, facilitating more efficient model aggregation.

For simplicity, we assume here that we are in the $i$th client.
Firstly, we initialize at each round the parameters of the teacher model $\bar{\param}_i$ and student model ${\param}_i$ with the parameters of the aggregated model $\param$.
During adaptation, the student model is updated through the minimization of the target loss function $\mathcal{L}_t(\cdot)$ which are later detailed in Sections \ref{sec:proto-server-training} and \ref{sec:local-training}.
After back-propagation through the student, we use the Exponential Moving Average (\textit{EMA}) parameters update \cite{Mean_Teachers, TempEns} to compute the teacher model.
At every iteration $t$, the parameters $\bar{\param}_i^{(t+1)}$ of the teacher model are given by:
\begin{equation}
    \label{eq:teacher}
   \bar{\param}_i^{(t+1)} = \tau \bar{\param}_i^{(t)} + (1-\tau) \param_i, 
\end{equation}
where $\tau \in [0,1)$ is a weighting factor.
The model $\hat{\param}_i$, which is sent back to the server for model aggregation, is assigned to the final teacher model $\bar{\param}_i^{(t)}$.

\subsection{Prototype estimation and server training}
\label{sec:proto-server-training}
\noindent \textbf{Source prototypes}. In the \textit{transmission stage}, the server sends prototypes to all the target clients. 
These prototypes are defined as the mean feature representation for each identity from the source domain.
Formally, the prototype $\boldsymbol{p}_k$ of the $k$th identity is given by:
\begin{equation}
    \boldsymbol{p}_k = \frac{1}{\vert \mathcal{S}_k \vert} \sum_{l \in \mathcal{S}_k} F_\param \left( \im_l^{\mathcal{S}} \right)~\forall~ 1 \leq k \leq K ,
\end{equation}
where $K$ is the number of identities in $\mathcal{S}$, $\mathcal{S}_k \subset \mathcal{S}$ is the set of images of the $k$th identity, and $\mathcal{S}_i \cap \mathcal{S}_j\!=\!\emptyset~\forall\!i\!\neq\!j$.
With enough diverse identities and images per identities from the source domain, the set of all source prototypes can serve as an approximation of the source domain distribution which can be transmitted with little cost. 
Subsequently, we use them to align the source and target distributions in the edge devices in the \textit{local training stage}.



\noindent \textbf{Pseudo-client loss}. The source domain is treated as a pseudo-client in the \textit{local-training stage}.
Since the pseudo-client has access to the source domain dataset with labeled samples $\mathcal{S}=\{(\im^{\mathcal{S}}_j, \vy^{\mathcal{S}}_j)\}_{j=1}^{N_s}$, we can compute a supervised source loss $\mathcal{L}_s$ for the $j$th sample as:
\begin{equation}
    \mathcal{L}_s(\im^{\mathcal{S}}_j,\vy^{\mathcal{S}}_j) = \mathcal{L}_{CEs} + \mathcal{L}_{Tris},
    \label{eqn:target_loss}
\end{equation}
with
\begin{gather}
    \nonumber
    \mathcal{L}_{CEs} = \beta_1 \mathcal{L}_{CE}\left( \mathcal{C}_s\circ F_{{\param}_s}(\im^{\mathcal{S}}_j), \vy^{\mathcal{S}}_j \right) + \beta_2 \mathcal{L}_{CE}\left(\mathcal{C}_s\circ F_{{\param}_s}(\im^{\mathcal{S}}_j), \mathcal{\bar C}_s\circ F_{\bar{\param}_s}(\im^{\mathcal{S}}_j) \right), \\
    \nonumber
    \mathcal{L}_{Tris} = \gamma_1 \mathcal{L}_{Tri}\left( F_{{\param}_s}(\im^{\mathcal{S}}_j), \vy^{\mathcal{S}}_j \right) + \gamma_2 \mathcal{L}_{Tri}\left( F_{{\param}_s}(\im^{\mathcal{S}}_j), F_{\bar{\param}_s}(\im^{\mathcal{S}}_j) \right),
\end{gather}
where $\mathcal{L}_{CE}$ is the cross-entropy loss, $\mathcal{\bar C}$ is the teacher classifier head, $\mathcal{L}_{Tri}$ is the triplet loss, $\beta_1 + \beta_2 = 1$, and $\gamma_1+\gamma_2 = 1$.

\subsection{Local training on edge devices}
\label{sec:local-training}

We now detail the \textit{local training stage} for the clients.
A key difficulty of the target domain training is the estimation of the number of identities from an unlabeled set of images $\mathcal{D}_i = \{\im^{(i)}_j \}_{j=1}^{N_i}$.
To this end, we apply the pseudo-labeling technique \cite{theory&practice,DGM,BottomUp} consisting of an iterative process between clustering with the DBSCAN \cite{dbscan} method and fine-tuning.
After this pseudo-labeling process we get an augmented dataset $\tilde{\mathcal{D}}_i = \{\im^{(i)}_j, \tilde{\vy}^{(i)}_j \}_{j=1}^{N_i}$, where $\tilde{\vy}^{(i)}_j$ is the pseudo-label associated to the $j$th sample.

\noindent \textbf{Target client loss}. 
In the edge devices, the teacher model generates soft labels that guide the student model to be less confident about the hard pseudo-labels \cite{SB}. 
This results in a refinement of the wrong predictions of the student model.
Specifically, for a given target client dataset $\mathcal{D}_i$, the local loss function $\mathcal{L}_i$ in a mini-batch is given by:
\begin{equation}
    \mathcal{L}_i = \frac{1}{m} \sum_{j \in \mathcal{D}_{i,m}} \mathcal{L}_p(\im^{(i)}_j) + \lambda \mathcal{L}_{MMD}\left(\mathcal{D}_{i,m},\mathcal{P}_m\right),
\end{equation}
where $\mathcal{D}_{i,m} \subseteq \mathcal{D}_{i}$ is the set of images in the mini-batch with $\vert \mathcal{D}_{i,m} \vert = m$, $\mathcal{L}_p(\im^{(i)}_j)$ is a pseudo-label loss for the $j$th sample, $\lambda$ is a weighting factor, and $\mathcal{L}_{MMD}\left(\mathcal{D}_{i,m},\mathcal{P}_m\right)$ is the MMD loss between $\mathcal{D}_{i,m}$ and a subset of the prototypes $\mathcal{P}_m \subseteq \{ \boldsymbol{p}_1, \boldsymbol{p}_2, \dots, \boldsymbol{p}_K \}$ with $\vert \mathcal{P}_m \vert = m$.
The pseudo-label loss $\mathcal{L}_p(\im^{(i)}_j)$ is the same as in \eqref{eqn:target_loss}, but since the true labels are not available in the clients, we use the pseudo-labels $\tilde{\vy}^{(i)}_j$ instead.
The local loss is used to update the student parameters $\param_i$.

\noindent \textbf{Personalized pseudo-epoch}. A significant challenge in federated learning scenarios is determining the optimal number of training epochs for each client. 
This decision is crucial to achieve the best balance between learning efficiency and transmission overhead.
In our task, this problem is also crucial to prevent over-fitting in clients with only a few identities or images.
To ensure equal usage of all identities within a client during a federated training round, we introduce the \textit{Personalized Pseudo-Epoch} (\textit{PPE}).

For a specific client \(i\), let \(K_i\) represent the count of identities in \(\mathcal{D}_i\) as identified by the DBSCAN algorithm. 
In every iteration, mini-batches are constructed by randomly selecting \(I\) identities. 
From each chosen identity, \(B\) images are sampled, as in previous works \cite{MMT,SpCL}.
Consequently, we define the number of iterations required for one PPE as \({P}_i = \frac{K_i}{I}\). 
By doing so, we ensure that, during a federated training round, each identity is presented an equal number of times, irrespective of the varying number of identities present in each client's dataset.

\section{Experiments and Results}
\label{sec:expes}

In this section, we detail our experimental setup, covering datasets, implementation details, and evaluation metrics.
Subsequently, we compare \method~against two categories of approaches: (i) FL + UDA, wherein we adapt the UDA methods MMT \cite{MMT} and SpCL \cite{SpCL} to \setting, and (ii) federated learning approaches for person Re-ID, namely FedReID \cite{FedReID} and FedUnReID \cite{FedUnReID}. 
Finally, we conduct a series of ablation studies to (i) demonstrate the efficacy of the transformer-based architecture coupled with self-supervised pre-training (\method++), (ii) confirm the suitability of the MMD loss, and (iii) validate the teacher-student architecture and aggregation choice.



\subsection{Experimental setup}

\noindent \textbf{Datasets}. We evaluate our method in real-to-real and synthetic-to-real scenarios.
For the source domain, we use two datasets:
\begin{itemize}
    \item \textit{MSMT} (MS) \cite{PTGAN} includes videos from $15$ cameras. 
    The training set has $32,621$ images of $1,042$ identities, while the test set comprises $11,659$ query images and $82,161$ gallery images from $3,060$ identities.
    \item \textit{RandPerson} (RP) \cite{RandPerson} is a synthetic dataset containing $8,000$ identities and $132,145$ images.
\end{itemize}

\noindent For the target domain, we consider the following datasets:
\begin{itemize}
    \item \textit{Market} (M) \cite{Market} has $1,501$ identities captured by six cameras. It includes $32,668$ images, with $12,936$ training images from $751$ identities and $19,732$ test images from the remaining $750$ identities.
    \item \textit{CUHK03-np} (C) \cite{cuhk03} comprises $14,097$ photos of $1,467$ individual identities where each identity is recorded by two cameras.
    We use the new protocol \cite{cuhk03-np} which consists of splitting the dataset into $767$ identities for training and $700$ identities for testing.
    In testing, each query identity is selected by both cameras to ensure the evaluation of the cross-camera Re-ID.
\end{itemize}

\noindent \textbf{Evaluation protocol}. In the \setting~setting, we assume the cameras are equipped with embedded devices that can train the teacher-student models of the clients.
To mimic this scenario, we split \textit{Market} into six clients and \textit{CUHK03-np} into two clients, where each client contains images from a single camera viewpoint.
We adopt the commonly used metrics for evaluation in person Re-ID \cite{MMT, SpCL}: mean Average Precision (mAP) and CMC Rank-1 \cite{Market} accuracies.
During each round of the federated learning, each client performs a number of PPEs.
Therefore, we compute the metrics on a separate test set related to the target domain using the aggregated model from the server.
We report for each method the highest average mAP and Rank-1, with the number of rounds required to reach these top scores.

\noindent \textbf{Implementation details}. For a fair comparison with the state-of-the-art methods, we follow the common practices in the UDA for person Re-ID field by adopting ResNet-50 \cite{resnet} pre-trained on ImageNet \cite{imagenet} as a backbone.
We train every method for $800$ rounds of federated learning.
Except for FedUnReID, where we follow its implementation details and set the training number of rounds to $200$.
We stop the training process upon observing any signs of divergence, specifically when there is a considerable decline in the test mAP over the training rounds.
We present a sensibility analysis of the hyper-parameters of \method~in the supplementary material.

To stress the practicality of the adopted setting, we also consider a variant of \method~called \method++, where we employ a stronger backbone architecture and leverage as initialization a model pre-trained on a large-scale Re-ID dataset. 
Concerning the architecture, we transition from the traditional ResNet-50 to a ViT \cite{TransReID} backbone.
We complement the backbone improvement with the adoption of self-supervised pre-trained models on the large-scale unlabeled dataset LUPerson \cite{LUP_self_sup}. 

\subsection{Comparison with the State of the Art}



Since \method~is the first method that addresses \setting, we adapt various methods initially designed for other settings to facilitate the comparison.

\noindent \textbf{Competitive methods}. 
We assess the performance of \method~against two federated frameworks for Re-ID: FedReID \cite{FedReID} and FedUnReID \cite{FedUnReID}.
On one hand, FedReID is a Fully Supervised (\textit{FS}) method that uses dynamic weight adjustment, knowledge distillation, and FedPav as its aggregation rule.
The original study of FedReID also explores our dataset partition in the edge devices, \ie, each client contains images from a single camera.
However, FedReID requires the target dataset $\mathcal{D}_i$ to be labeled, while we do not have such a constraint.
On the other hand, FedUnReID is a framework that adapts the Purely Unsupervised (\textit{PU}) baseline Bottum-Up-Clustering (\textit{BUC}) \cite{BottomUp} for Federated person Re-ID.
We also compare \method~with FedReID+$\mathcal{S}$ and FedUnReID+$\mathcal{S}$.
These variants are improved versions of the original frameworks where we initialize the models with supervised source pre-training, offering a fairer comparison with \method~that leverages the source domain's knowledge.

Since our \setting~setting combines both UDA and FL, we extend our comparison to include \method~against UDA methods for person Re-ID.
For the UDA methods, we adapt the state-of-the-art pseudo-labeling approaches SpCL and MMT to suit the \setting~setting.
In this process, during each federated learning round, we send copies of these UDA frameworks to all the edge clients for local training.
Additionally, for a fair comparison with \method, we train in the server the pseudo-source client on the labeled source domain $\mathcal{S}$.
The aggregation rule for these adapted UDA methods, denoted FedAvg+SpCL and FedPav+MMT is consistent with Eq.~\eqref{aggregation_rule}.
The main objective of this comparison is to evaluate the effectiveness of traditional UDA methods when confronted with privacy constraints, where the target domain is distributed over multiple edge devices (cameras).

\noindent \textbf{Quantitative results and discussions}. Table \ref{tab:main_tab} reports the best mAP accuracy and CMC Rank-1 score alongside the number of rounds (\#R) required to achieve these top scores.
We include two real-to-real configurations MS $\to$ M, MS $\to$ C, and two synthetic-to-real configurations RP $\to$ M, RP $\to$ C.

\method~demonstrates good results against the supervised and unsupervised federated learning methods for person Re-ID, FedReID, and FedUnReID as shown in Table \ref{tab:main_tab}.
For example, \method~obtains $23.8$ of mAP in MS $\to$ C, outperforming FedReID with $11.6$ mAP and FedUnReID with $6.8$ mAP.
More interestingly, \method~reaches this performance after only $22$ rounds, whereas FedReID and FedUnReID require $750$ and $170$ rounds, respectively.
\method~also reaches superior performance in the RP $\to$ C configuration with $25.1$ mAP compared to the other federated learning methods.
We also evaluate the improved versions FedReID+$\mathcal{S}$ and FedUnReID+$\mathcal{S}$ where both models start with a pre-training on the source domain.
Even though starting from the source pre-trained models improves slightly the original models' performances, they are below the performances obtained by \method~in almost all configurations.

\begin{table}[tb]
    \centering
    \caption{Comparison of mAP, Rank-1 accuracy, and number of rounds (\#R) for four adaptation configurations. The different methods range from Fully Supervised (FS) and Purely Unsupervised (PU) to Unsupervised Domain Adaptation (UDA). $^*$The communication cost for a single round in MMT is four times greater than that in the other ResNet-based models.}
    \resizebox{\textwidth}{!}{
    \begin{tabular}{lccccccccccccc}
    \toprule
    \multirow{2}{*}{\textbf{Method}} & \multirow{2}{*}{\textbf{Type}} & \multicolumn{3}{c}{\textbf{MS} $\to$ \textbf{M}} & \multicolumn{3}{c}{\textbf{MS} $\to$ \textbf{C}} & \multicolumn{3}{c}{\textbf{RP} $\to$ \textbf{M}} & \multicolumn{3}{c}{\textbf{RP} $\to$ \textbf{C}} \\ 
    & & \textbf{mAP} & \textbf{Rank-1} & \textbf{\#R} & \textbf{mAP} & \textbf{Rank-1} & \textbf{\#R} & \textbf{mAP} & \textbf{Rank-1} & \textbf{\#R} & \textbf{mAP} & \textbf{Rank-1} & \textbf{\#R} \\
    \midrule
    FedReID \cite{FedReID}& FS & $38.9$ & $61.9$ & $800$ & $11.6$ & $11.7$ & $750$ & $38.9$ & $61.9$ & $800$ & $11.6$ & $11.7$ & $750$ \\
    FedReID+$\mathcal{S}$ & FS & $39.5$ & $63.8$ & $790$ & $12.0$ & $12.3$ & $800$ & \underline{$40.0$} & $64.4$ & $800$ & $11.4$ & $11.6$ & $780$ \\  \cdashline{1-14}
   
    FedUnReID \cite{FedUnReID} & PU & $19.5$ & $43.6$ & $190$ & $6.8$ & $7.0$ & $170$ & $19.5$ & $43.6$ & $190$ & $6.8$ & $7.0$ & $170$ \\
    FedUnReID+$\mathcal{S}$ & PU & $31.0$ & $61.7$ & $170$ & $10.5$ & $11.1$ & $170$ & $31.5$ & $31.8$ & $170$ & $10.6$ & $11.6$ & $160$ \\  \cdashline{1-14}
    FedAvg+SpCL \cite{SpCL} & UDA & $39.1$ & $67.3$ & $8$ & $19.7$ & $18.9$ & $1$ & $36.1$ & $62.9$ & $9$ & $21.2$ & $21.6$ & $3$  \\
    FedPav+MMT$^*$ \cite{MMT}& UDA & $45.8$ & $73.6$ & $70$ & $22.4$ & $21.9$ & $9$ & $30.2$ & $58.9$ & $9$ & $19.0$ & $19.7$ & $9$ \\
    \method~(\textcolor{red}{ours})& UDA & \underline{$51.0$} & \underline{$76.8$} & $288$ & \underline{$23.8$} & \underline{$23.1$} & $22$ & $39.2$ & \underline{$66.4$} & $22$ & \underline{$25.1$} & \underline{$24.7$} & $253$ \\
    \method++ (\textcolor{red}{ours})& UDA & {$\textbf{61.7}$} & {$\textbf{82.6}$} & $170$ & {$\textbf{43.8}$} & {$\textbf{42.4}$} & $24$ & {$\textbf{45.2}$} & {$\textbf{71.8}$} & $186$ & {$\textbf{25.7}$} & {$\textbf{24.9}$} & $212$ \\
    \bottomrule
    \end{tabular}
}
    \label{tab:main_tab}
\end{table}

\method~also improves significantly the performance of the adapted UDA baselines SpCL and MMT as shown in Table \ref{tab:main_tab}.
In MS $\to$ M, SpCL and MMT achieve a mAP of $39.1$ and $45.8$, respectively, while \method~achieves a mAP accuracy of $51$.
This observation can also be generalized to the synthetic-to-real configurations like RP $\to$ M, where \method~reaches a performance of $39.2$ mAP, while SpCL and MMT achieve $36.1$ and $30.2$ mAP, respectively.
Even though \method~requires more communication rounds to reach its optimal performance compared to MMT, it is important to notice that \method~transmits approximately only a quarter of the data weights per round. 
This is because the MMT architecture sends four backbones to the server, whereas \method~needs to share only one (the teacher model) and the transmission cost of the prototypes is almost negligible compared to the weights of the models.
Overall, \method~is more effective in the \setting~scenario than the adapted UDA baselines SpCL and MMT as shown in Table \ref{tab:main_tab}.

\noindent \textbf{\method++}
Recent work \cite{LUP_self_sup} has shown the suitability and effectiveness of self-supervised pre-training methods for transformer-based methods \cite{TransReID} in person Re-ID, yielding substantial enhancements across a variety of Re-ID benchmarks.
In the context of our \setting~setting, the performance of \method++ is consistent with the aforementioned findings as shown in Table \ref{tab:main_tab}.
Particularly, transitioning from the ResNet-50 to a ViT backbone pre-trained in a self-supervised way leads to remarkable performance enhancements in all the configurations.
For instance, we observe an increase in the mAP from $51$ to $61.7$ in MS $\to$ M.
Similarly, we have an improvement from $39.2$ to $45.2$ in RP $\to$ M in the mAP, showing \method++ enhanced effectiveness.
The improvement in the performance of using transformer-based models in person Re-ID comes from three main reasons \cite{TransReID}: (i) the multi-head self-attention effectively captures long-range dependencies and drives the model to focus on diverse human-body parts, (ii) transformer-based models have the ability to extract fine-grained features which is essential in person Re-ID, and (iii) the rich variety and volume of the LUPerson dataset provide the model with the capability of extracting more robust features that are generalizable across small downstream datasets.
We perform an ablation study in Section \ref{sec:abl_studies} to empirically validate these points.

\begin{figure}[t]
    \centering

    \begin{minipage}{0.53\textwidth}
        \centering
        \setlength\tabcolsep{2pt} 
        \captionof{table}{Impact of the backbone architecture and pre-training datasets on \method.}
        \vspace{0.4cm}
        \resizebox{0.95\textwidth}{!}{
        \begin{tabular}{ccccc}
            \toprule
            \textbf{Backbone} & \textbf{Pre-tr.} & \textbf{Warm-up} & \textbf{MS} $\to$ \textbf{M} & \textbf{MS} $\to$ \textbf{C} \\
            \midrule
            ResNet-50 & ImageNet & \ding{55} & $41.5$ & $23.7$ \\
            ResNet-50 & ImageNet & \ding{51} & $\textbf{51.0}$ & $\textbf{23.8}$ \\
            ResNet-50 & LUPerson & \ding{55} & $44.0$ & $13.6$ \\
            ResNet-50 & LUPerson & \ding{51} & $46.0$ & $16.0$ \\
            \cdashline{1-5}
            ViT (S) & ImageNet & \ding{51} & $52.4$ & $27.5$ \\
            ViT (S) & LUPerson & \ding{55} & $59.7$ & $23.9$ \\
            ViT (S) & LUPerson & \ding{51} & {$\textbf{61.7}$} & {$\textbf{43.8}$} \\
            \bottomrule
        \end{tabular}
        }
        \label{tab:transformers}
    \end{minipage}\hfill
    \begin{minipage}{0.47\textwidth}
        \includegraphics[width=\textwidth]{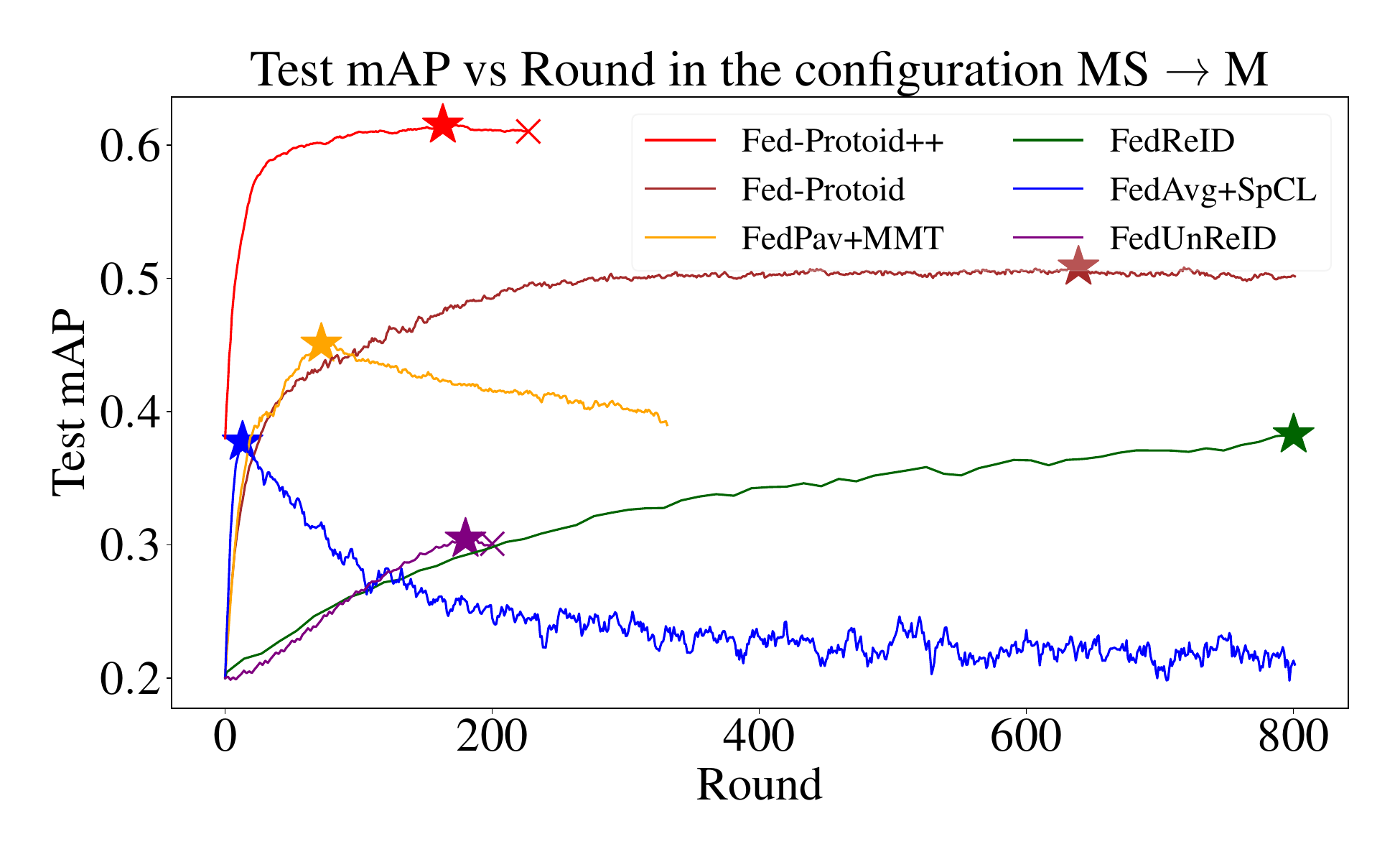}
        
        \vspace{-0.5cm}
        \caption{Test mAP vs Round (MS $\to$ M). \raisebox{-.3ex}{\includegraphics[height=0.3cm]{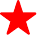}}~denotes the maximum mAP.}
        \label{fig:results}
    \end{minipage}
\end{figure}

\noindent \textbf{Training dynamics}.
In Fig. \ref{fig:results}, we illustrate the progression of the mAP of the different methods in the MS $\to$ M configuration.
Notably, there is a difference in the evolution of the mAP between the methods designed for the FL FedReID+$\mathcal{S}$ and FedUnReID+$\mathcal{S}$, and the UDA-based methods FedAvg+SpCL and FedPav+MMT.
Specifically, while FedReID+$\mathcal{S}$ and FedUnReID+$\mathcal{S}$ exhibit a consistent improvement during the training, this trend is not mirrored in the performance of FedAvg+SpCL and FedPav+MMT.
In fact, both UDA-based methods tend to converge rapidly at the early stage of FL training.
This is because initially, the local models are relatively close to the source model, allowing for easier leveraging of the source domain knowledge in the first rounds of FL.
However, as training progresses, the local models start diverging from the source model, leading to a decrease in performance.
Conversely, our methods demonstrate a stable progression, effectively managing to mitigate domain shift during training. 
This highlights the effectiveness of our approach in maintaining consistent performance in the \setting~setting.

\subsection{Ablation studies}
\label{sec:abl_studies}



\noindent \textbf{On the effectiveness of the backbones and pre-training datasets}. The first ablation study focuses on the impact of the different modifications done to design \method++.
Table \ref{tab:transformers} shows the ablation study for different backbones, pre-training strategies, and warm-up.
For the backbones, we have the option to use either the classical ResNet-50 or ViT Small (S).
For the pre-training dataset and strategy, we can either use fully supervised on ImageNet or self-supervised in LUPerson.
The warm-up consists of adding an additional supervised pre-training on the source domain $\mathcal{S}$.
We observe in Table \ref{tab:transformers} that adopting the ViT backbone combined with an appropriate pre-training dataset significantly enhances the performance.
\method~corresponds to a ResNet-50 backbone pre-trained on ImageNet, while \method++ corresponds to a ViT model pre-trained on the large-scale LUPerson dataset in a self-supervised way.


A key finding in Table \ref{tab:transformers} is that using ViT (S) instead of ResNet-50 with the same pre-training strategy consistently results in performance improvements.
For instance, when comparing ResNet-50 and ViT (S) pre-trained on ImageNet, the mAP slightly improves from $51$ to $52.4$ in MS $\to$ M, and from $23.8$ to $27.5$ in MS $\to$ C.
This suggests that the ViT-based backbone learns more robust features in the target domain.
Additionally, ViT (S) captures more generalizable features when pre-trained in a self-supervised way thanks to the large and diverse set of unlabeled images in LUPerson.
As a final remark, the warm-up generally enhances the performances across all the scenarios and configurations.

\noindent \textbf{On the effectiveness of the teacher-student framework}.
The integration of the teacher-student architecture gives multiple possibilities to the design of \method.
Table \ref{tab:teacher_student} shows the results of the ablation study where (i) we do not have the teacher-student framework in the pseudo-client, (ii) we have the teacher-student framework in the pseudo-client and we transmit the students for aggregation, and (iii) we have the teacher-student framework in the pseudo-client and we transmit the teachers for aggregation.
Table \ref{tab:teacher_student} shows a considerable drop in performance when the teacher model is omitted from the pseudo-client in both MS $\to$ M and MS $\to$ C.
Specifically, the mAP decreases from $51$ to $37.4$ in  MS $\to$ M, and from $23.8$ to $22.5$ in MS $\to$ C, underscoring the crucial role of the teacher model in the pseudo-client.
These findings align with our claim in Section \ref{sec:teacher-student} regarding the use of the teacher-student architecture in the pseudo-client to keep similar training dynamics with the other clients.
The teacher-student architecture gives another alternative of transmitting the student instead of the teacher models for aggregation.
Table \ref{tab:teacher_student} illustrates that this alternative yields reasonable performance.
However, using students for aggregation falls marginally short of the performance achieved by aggregating the teacher models.

\begin{table}[t]
\parbox{.5\linewidth}{
\centering
    \caption{Ablation study of the teacher-student framework and the choice of the transmitted model.}
    \resizebox{0.5\columnwidth}{!}{\begin{tabular}{cccc}
    \toprule
       \textbf{Teach.-Stud.} & \multirow{2}{*}{\textbf{Transmission}} & \multirow{2}{*}{\textbf{MS} $\to$ \textbf{M}} & \multirow{2}{*}{\textbf{MS} $\to$ \textbf{C}} \\
       \textbf{on} $\mathcal{S}$ &  &  & \\
       \midrule
     \ding{55} & Teacher & $37.4$ & $22.5$ \\
     \ding{51} & Student & \underline{$49.3$} & $\textbf{23.8}$ \\
     \ding{51} & Teacher & $\textbf{51.0}$  & $\textbf{23.8}$ \\
     \bottomrule
    \end{tabular}}
    \label{tab:teacher_student}
}
\hfill
\parbox{.48\linewidth}{
\centering
   \caption{Impact of the kernel function choice for the Maximum Mean Discrepancy (MMD) loss.}
   \resizebox{0.40\columnwidth}{!}{ \begin{tabular}{cccc}
        \toprule
        \textbf{MMD} & \textbf{Kernel} & \textbf{MS} $\to$ \textbf{M} & \textbf{MS} $\to$ \textbf{C} \\
         \midrule
         \ding{55} & -- & 42.6 & 22.4\\
         \ding{51} & Linear & 38.1 & 22.1\\
         \ding{51} & Order 2 & 27.8 & 13.1\\
        \ding{51} & Gaussian &\textbf{51.0}  & \textbf{23.8} \\  \bottomrule
    \end{tabular}    }
    \label{tab:mmd_loss}
}
\end{table}


\noindent \textbf{On the effectiveness of the MMD loss}.
The MMD loss, serving as a measure of domain discrepancy, offers a variety of options for the reproducing kernel Hilbert space where we minimize the distance between source prototypes and target feature representations.
Table \ref{tab:mmd_loss} shows a comparison between different kernel functions including linear, order 2, and Gaussian kernels.
The linear kernel minimizes the mean average of prototypes and target features distributions, while the order 2 kernel minimizes the mean average and the standard deviation of these distributions.
Table \ref{tab:mmd_loss} suggests that the linear and order 2 kernels are not effective in the \setting~setting.
This can be attributed to potentially biased estimations of the true mean (linear) and variance (2nd order) within relatively small and diverse batches of images.
Furthermore, using the MMD loss with a Gaussian kernel achieves superior performance in all cases, including when MMD loss is not used at all. 
We further evaluate the MMD's effectiveness by examining its performance with limited prototypes and comparing the proposed distributed MMD with the original MMD in the supplementary material, demonstrating its robustness against device storage and communication limitations and proving it to be effective and suitable for our setting.
\section{Conclusion}

In this paper, we presented a novel approach for the task of UDA for person Re-ID that addresses both problems of domain shift and privacy preservation.
Our method \method~ learns a person Re-ID model across multiple edge devices without transmitting target images from the cameras where they were captured.
By integrating a teacher-student architecture and a source-client model, trained in the server side on labeled source domain, and introducing a distributed version of the Maximum Mean Discrepancy (MMD) loss, \method~ensures effective domain adaptation with the target clients while eliminating the need for explicit inter-camera learning and keeping communication requirements minimal.

%
%
\bibliographystyle{splncs04}
\bibliography{main}

\def\method{Fed-Protoid} 
\def\setting{DUDA-Rid}


%





\title{Privacy-Preserving Adaptive Re-Identification without Image Transfer :
Supplementary Materials}

\titlerunning{Fed-Protoid}

\author{Hamza Rami\inst{1,2}\orcidlink{0009-0002-3654-3878} \and
Jhony H. Giraldo\inst{1}\orcidlink{0000-0002-0039-1270} \and
Nicolas Winckler\inst{2}\and
Stéphane Lathuilière\inst{1}}

\authorrunning{H. Rami et al.}

\institute{LTCI, Télécom Paris, Institut Polytechnique de Paris \and Atos}

\maketitle
In this supplementary material, we provide results and analysis of additional experiments and present additional details about the Fed-Protoid.
\begin{itemize}
    \item We provide the pseudo-code of Fed-Protoid to give more details about its algorithmic structure.
    \item We present a set of experiments focused on the source prototypes, initially showing the significance of utilizing the global model for their computation instead of the pseudo client. Subsequently, we highlight the impact of reducing the number of source prototypes in the performance of \method.
    \item We include a detailed analysis regarding the sensibility of the hyper-parameters of \method.
    \item We compare the distributed MMD with the original MMD and some Domain Generalization (DG) Re-ID methods.
\end{itemize}

\section{Fed-Protoid: Algorithm}

For completeness, we detail the Fed-Protoid algorithm as follows:

\begin{algorithm}
\caption{Fed-Protoid algorithm}
\begin{algorithmic}[1]
\State \textbf{Input:} $n$ unlabeled datasets $\{\mathcal{D}_1, \mathcal{D}_2, \dots, \mathcal{D}_n\}$, an annotated dataset $\mathcal{S}$, and the source pre-trained weights $\param_s$

\State Initialize model $F_\param$ with parameters $\param_s$
\For{each training round}
    \State \textbf{Transmission Stage:}
    \State Transmit $F_\param$ and source prototypes to all clients
    \State \textbf{Local Training Stage:}
    \State Update pseudo-client model $F_{\hat{\param}_s}$ using $\mathcal{S}$ and $\mathcal{L}_s$
    \For{each client $i$}
        \State Update client model $F_{\hat{\param}_i}$ using local dataset $\mathcal{D}_i$ and $\mathcal{L}_i$
    \EndFor

    \State \textbf{Aggregation Stage:}
    \State Aggregate models using equation $\param = \alpha \hat{\param}_s + (1 - \alpha) \sum_{i=1}^{n} w_i \hat{\param}_i$

\EndFor

\State \textbf{Output:} Trained federated model $F_\param$
\end{algorithmic}
\end{algorithm}

\section{Additional experiments on the source prototypes: computation and communication}

\subsection{Impact of Global model in source prototype computation}
In this section, we present experimental results for both \method~and \method++, showing the advantages of utilizing the global model for computing source prototypes.
The results shown in Table \ref{tab:pseudo_client_protos} indicate that across the two configurations MS $\to$ M and MS $\to$ C, we obtain superior performance when the prototypes are derived from the global model instead of the pseudo-client model.
The effectiveness of using the global model in prototype computation can be attributed to its ability to bridge the gap between the source and target domain distributions.
Essentially, the prototypes generated by the global model represent a median distribution that lies between those of the source and target domains. 
This intermediary positioning facilitates more efficient optimization of the Maximum Mean Discrepancy (MMD). 
Instead of directly aligning the target domain with the source domain, the global model provides features that are equidistant to both domains. 
Consequently, this approach converges all distributions towards a central, unified distribution, rather than skewing them towards the source domain distribution alone.

\begin{table}
    \centering
    \caption{Ablation study of the choice of the model that computes the source prototypes.}
    \begin{tabular}{cccc}
      \toprule
         \multirow{2}{*}{\textbf{Method}} & {\textbf{Source Prototypes}} & \multirow{2}{*}{\textbf{MS} $\to$ \textbf{M}} & \multirow{2}{*}{\textbf{MS} $\to$ \textbf{C}} \\
        & \textbf{computed with} &  & \\
       \midrule
      \multirow{2}{*}{Fed-Protoid} & Pseudo-client & $43.1$ & $23.6$ \\ 
     & Global model & $\textbf{51.0}$ & $\textbf{23.8}$ \\ 
     \cdashline{1-4}
     \multirow{2}{*}{Fed-Protoid++}& Pseudo-client & $60.5$ & $43.7$ \\
     &Global model & $\textbf{61.7}$ & $\textbf{43.8}$ \\
     \bottomrule
    \end{tabular}
    
    \label{tab:pseudo_client_protos}
\end{table}
\subsection{The impact of the number of source prototypes}
\begin{figure}[h]
 \def\myim#1{ \includegraphics[width=.72\columnwidth]{#1}}
     \centering
   \setlength\tabcolsep{-0.0 pt}
   \renewcommand{\arraystretch}{0.01}
     \vspace{-0.37cm}\begin{tabular}{ccc}
\makecell[l]{\myim{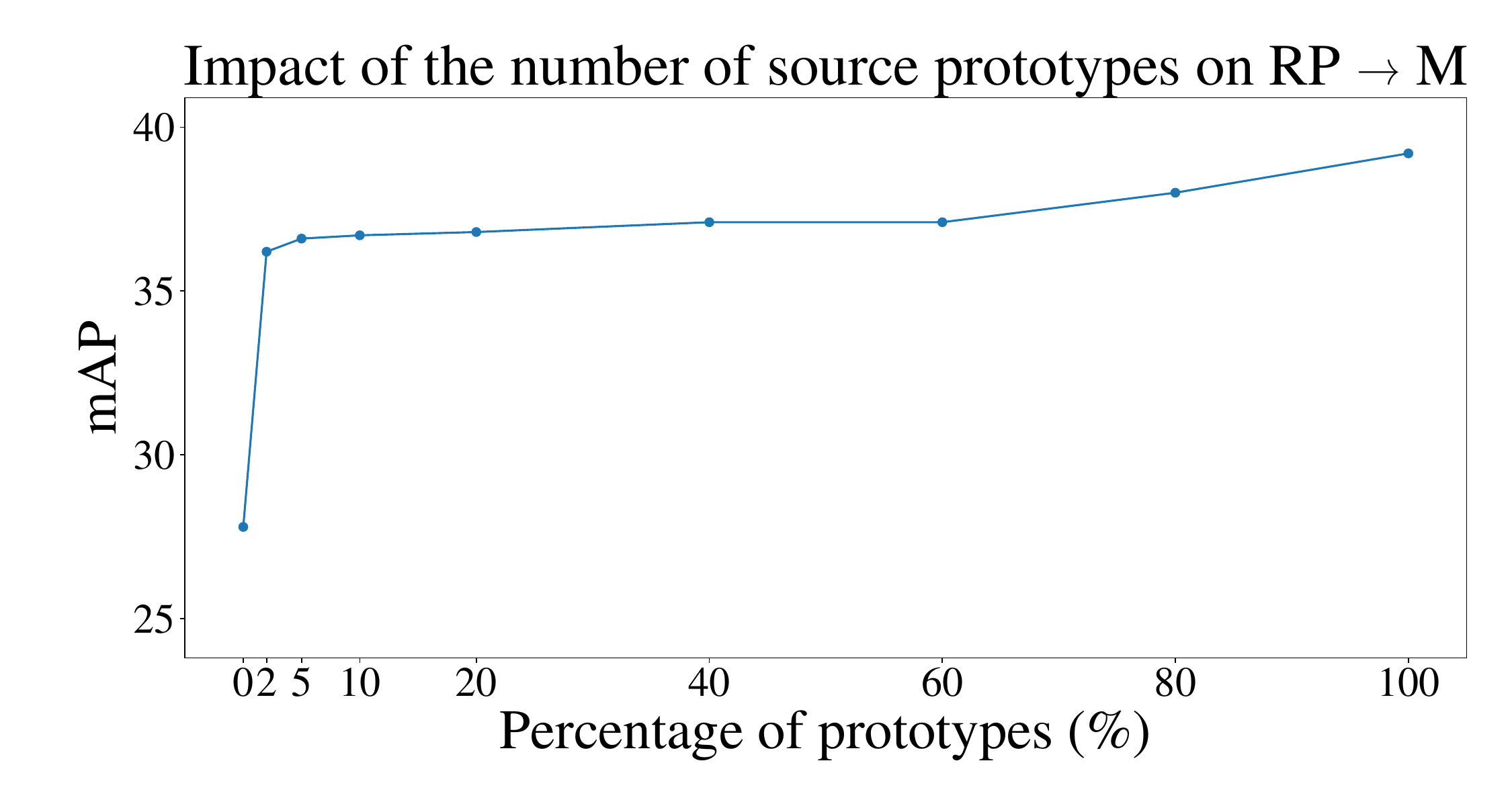}} & \\
\makecell[l]{\myim{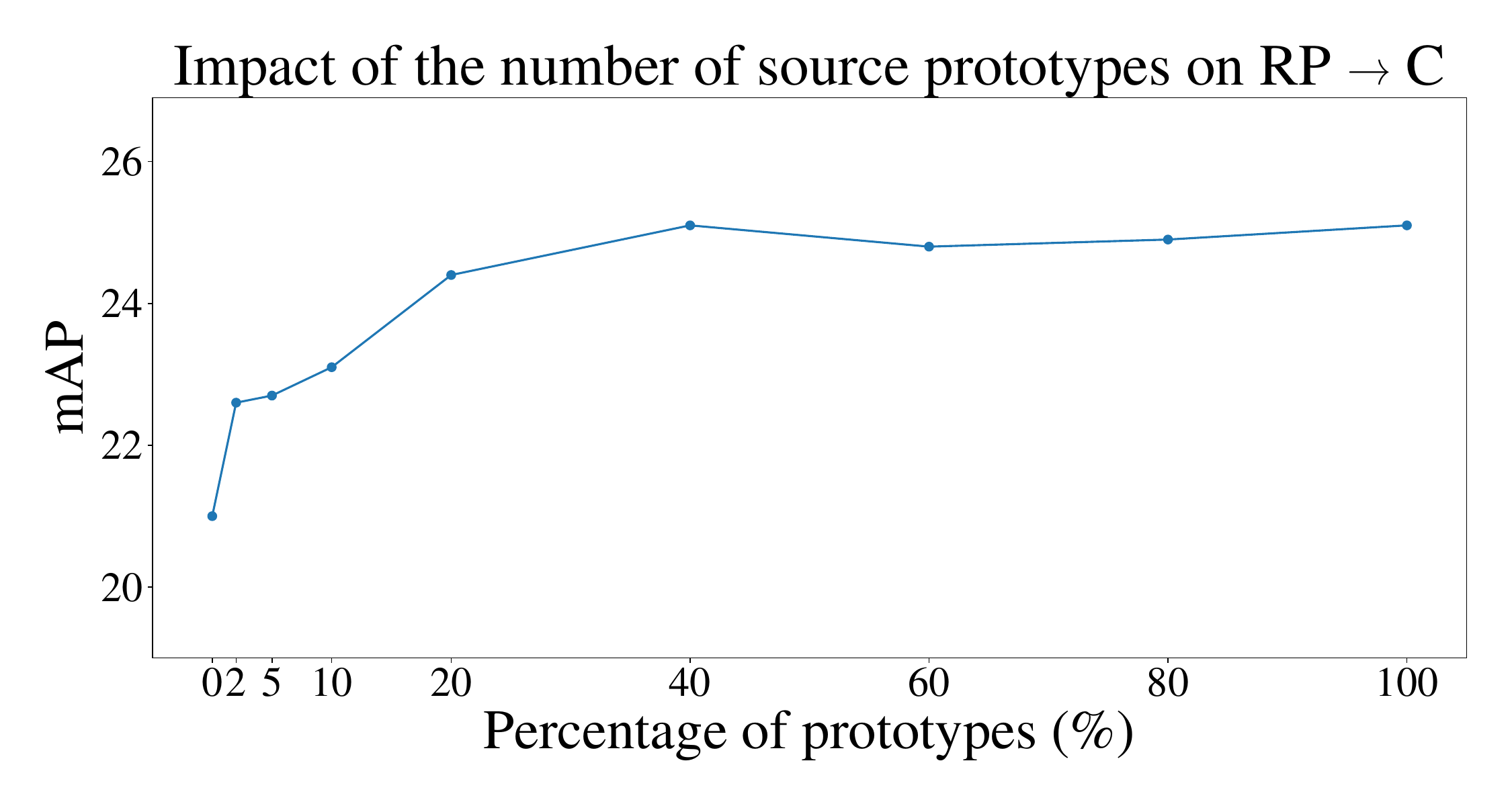}}
      \end{tabular}
\caption{The impact of the number of source prototypes in the \method~performance in two configurations: RP$\to$M and RP$\to$C}
\label{fig:percent}
\end{figure}

The following experiments aim to assess the effectiveness of \method~in scenarios with stronger memory and communication limitations.
Such situations occur when a large source domain with many identities is deployed in the server, resulting in an increased number of prototypes, thus requiring higher communication bandwidth.
For instance, the synthetic RP dataset contains $8,000$ identities which is $8$ times the number of identities in the MS dataset. To address this challenging scenario, we propose investigating whether a simple uniform sub-sampling of prototypes reduces the transmission cost without impacting the ReID performance.
We evaluate \method~with a reduced number of source prototypes in both configurations: RP$\to$M and RP$\to$C.
Fig. \ref{fig:percent} shows that \method~remains effective despite a significant decrease in the number of source prototypes for the MMD optimization on edge devices.
\method~shows a stable performance when the number of source prototypes varies between $2\%$ and $100\%$ in RP$\to$M and between $20\%$ and $10\%$ in RP$\to$C configuration.
Therefore, we argue that \method~is robust against device storage and communication limitations.
\section{Variability of \method~and hyper-parameters}
\subsection{Variability of the performance of Fed-Protoid across different initialization}

We conduct multiple experiments of our \method~and \method++ with three different seeds in all the configurations presented in the main paper. 
We report in Table \ref{tab:variance_runs} the mean of the mAP and Rank-1 across those runs.
Alongside these metrics, we also report the standard deviation to illustrate the variability in the results.
We can state that we have consistency and minimal variance across all the different configurations, demonstrating the robustness and reliability of our method under different initializations.

\begin{table}[h]
    \centering
    \caption{Standard deviation of both Fed-Protoid and Fed-Protoid++ with varying seeds.}
    \begin{tabular}{ccccc}
    \toprule
    \multirow{2}{*}{\textbf{Configuration}} & \multicolumn{2}{c}{\textbf{Fed-Protoid}} & \multicolumn{2}{c}{\textbf{Fed-Protoid++}} \\
     & \textbf{mAP} & \textbf{Rank-1} & \textbf{mAP} & \textbf{Rank-1} \\
   \midrule
     MS $\to$ M & $51.0_{\pm \num{0.3}}$ & $76.8_{\pm \num{0.2}}$ & $61.7_{\pm \num{0.2}}$ & $82.6_{\pm \num{0.1}}$ \\
     MS $\to$ C & ${23.8_{\pm \num{0.2}}}$ & ${23.1_{\pm \num{0.1}}}$ & $43.8_{\pm \num{0.2}}$ & $42.4_{\pm \num{0.4}}$ \\
     RP $\to$ M & ${39.2}_{\pm \num{0.1}}$ & ${66.4}_{\pm \num{0.0}}$ & $45.2_{\pm \num{0.1}}$ & $71.8_{\pm \num{0.1}}$\\
     RP $\to$ C & $25.1_{\pm \num{0.4}}$ & $24.7_{\pm \num{0.3}}$ & $25.7_{\pm \num{0.2}}$ & $24.9_{\pm \num{0.1}}$ \\
     \bottomrule
    \end{tabular}
    
    \label{tab:variance_runs}
\end{table}

\subsection{Hyper-parameters ablation study} 

\begin{figure}
 \def\myim#1{ \includegraphics[width=.34\columnwidth]{#1}}
     \centering
   \setlength\tabcolsep{-0.0 pt}
   \renewcommand{\arraystretch}{0.01}
     \vspace{-0.37cm}\begin{tabular}{ccc}
\makecell[l]{\myim{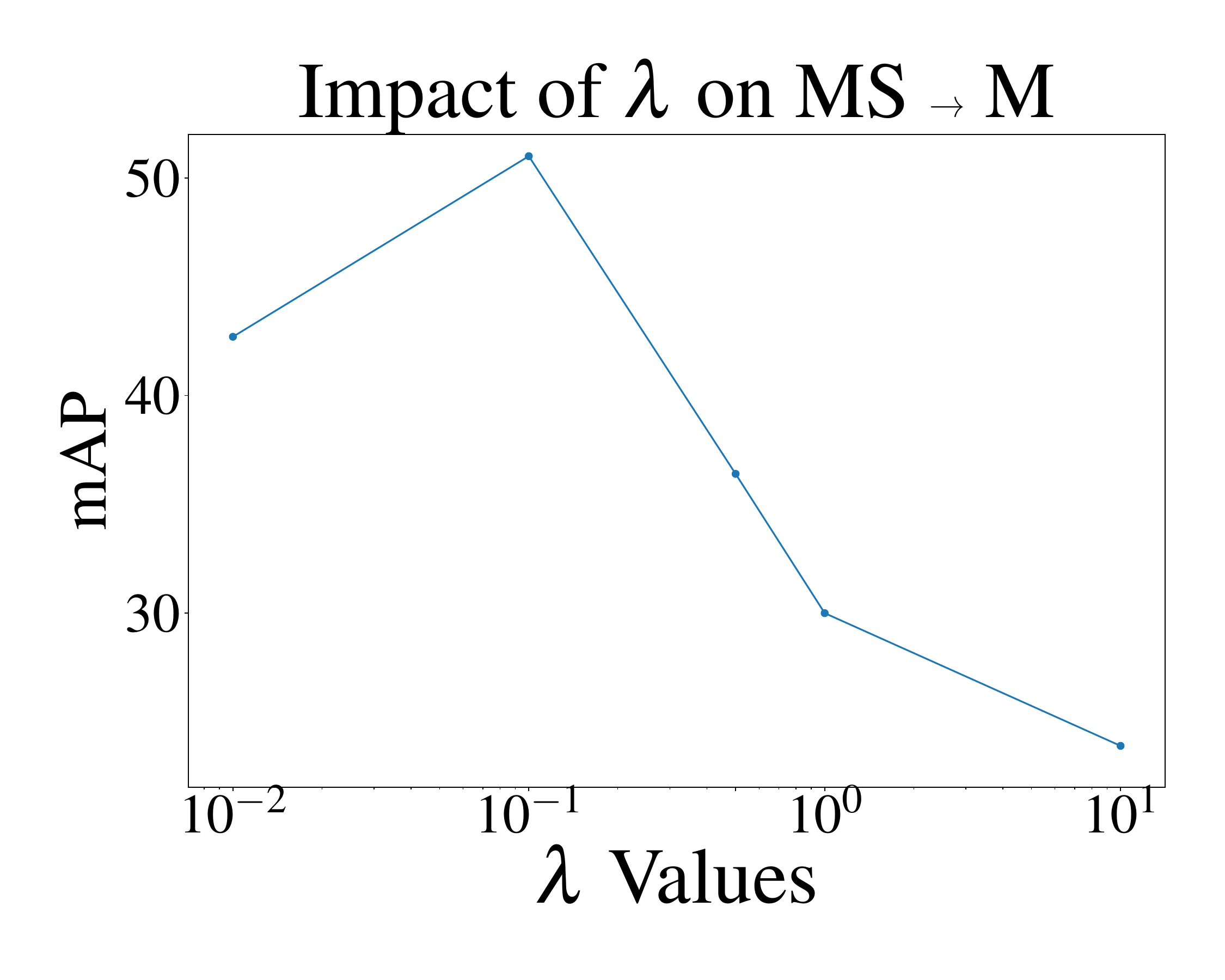}} &
\makecell[l]{\myim{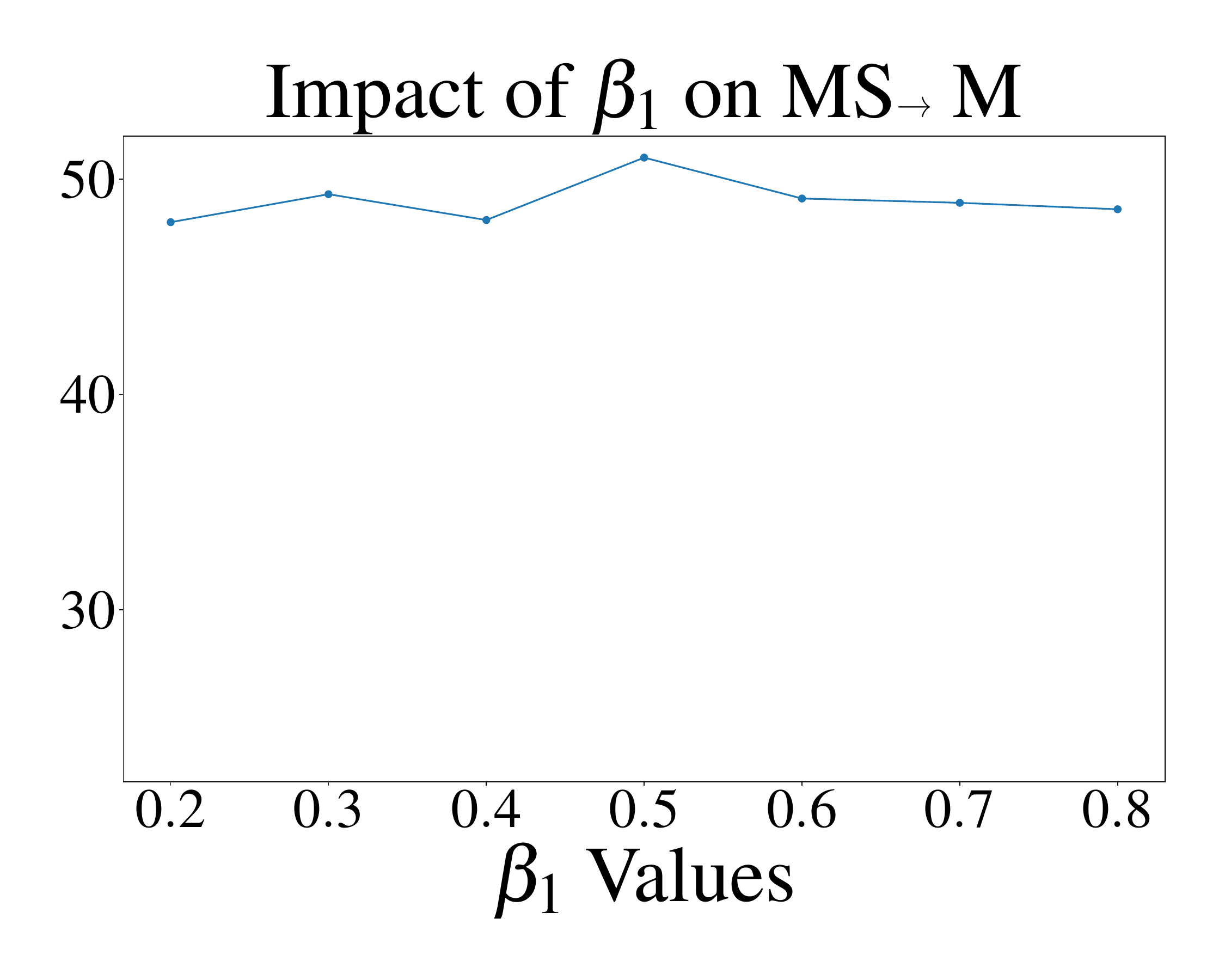}} &
\makecell[l]{\myim{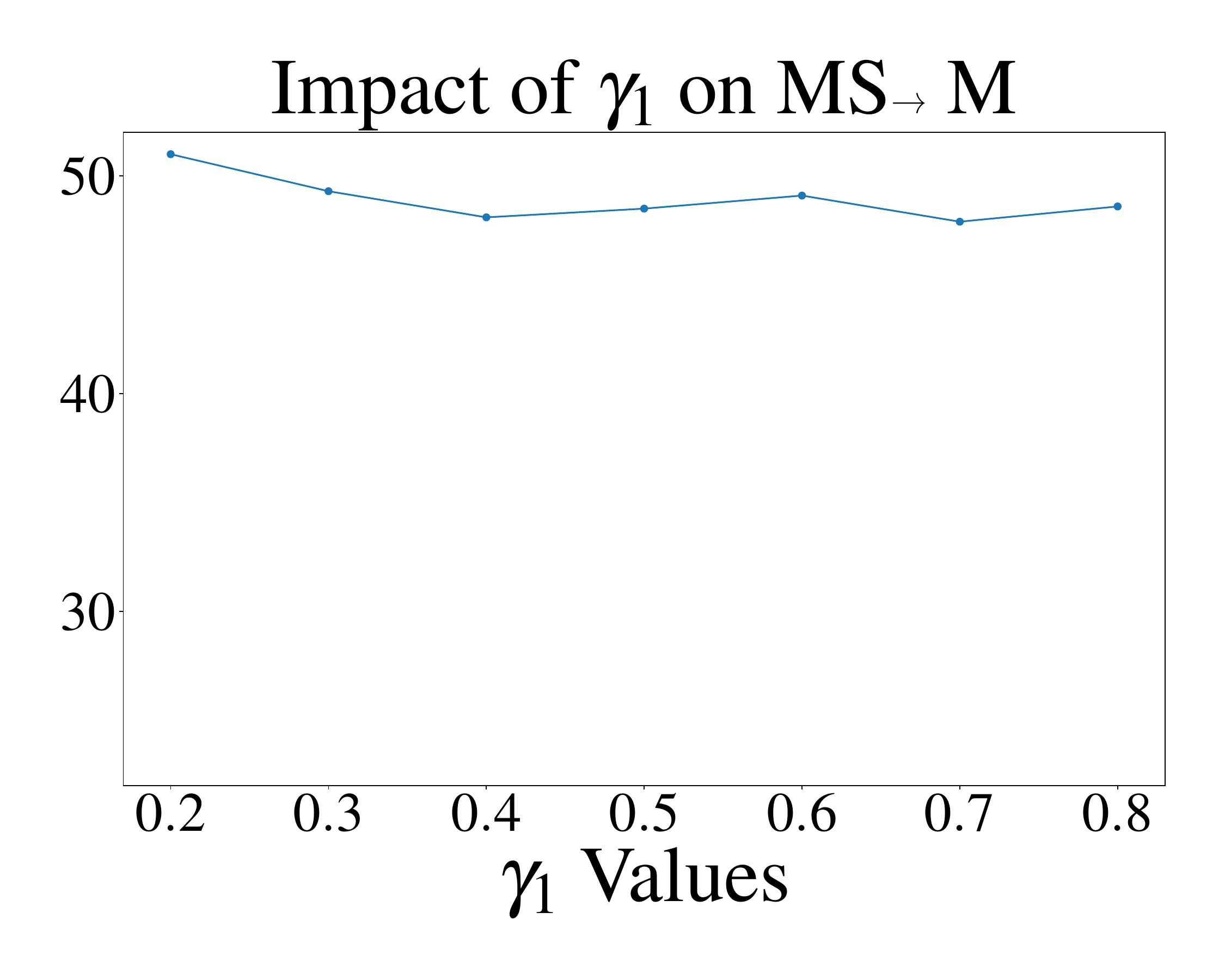}}
      \end{tabular}
\caption{Ablation study on the sensibility of the different hyper-parameters of \method.}
\label{fig:hyperparameters}
\end{figure}

Fig. \ref{fig:hyperparameters} illustrates the results of the sensitivity analysis conducted on the hyper-parameters of \method.
In this analysis, all hyper-parameters except the specific one under investigation are maintained at their default values.
We observe that changing the hyper-parameters $\beta_1$ and $\gamma_1$ results in a slight impact on the accuracy with only minimal variances, showing that our method is stable and robust.
As for $\lambda$, which is the hyper-parameter controlling the importance of the MMD loss in the final objective, we determined that a value of $0.1$ yields optimal results and have therefore set it to this fixed value for subsequent experiments.

\clearpage

\section{Distributed MMD vs. Original MMD}
Tab. \ref{tab:main_tab_reb_mmd} compares the distributed MMD with the original MMD. 
For the original MMD, we use the same DUDA-Rid setting, but the MMD loss is computed over the entire target dataset.
The results show that the distributed MMD outperforms the original MMD in both configurations.
Note that the original MMD violates DUDA-Rid privacy constraints, making it unsuitable for our problem.

\begin{table}[h]
    \centering
    \vspace{-18pt}
    \caption{Comparison between original and distributed MMD.}
    \resizebox{0.8\columnwidth}{!}{
    \begin{tabular}{lcccc}
    \toprule
    \multirow{2}{*}{\textbf{Method}}  & \multicolumn{2}{c}{\textbf{MS} $\to$ \textbf{M}} & \multicolumn{2}{c}{\textbf{RP} $\to$ \textbf{M}} \\ 
     & \textbf{mAP} & \textbf{Rank-1}    & \textbf{mAP} & \textbf{Rank-1}   \\
    \midrule
    Fed-Protoid + orig. MMD & {$47.1$} & {$75.3$}    & $30.2$ & {$59.2$}   \\
    Fed-Protoid~+ dist. MMD (\textcolor{red}{ours}) & {$\textbf{51.0}$} & {$\textbf{76.8}$} & $\textbf{39.2}$ & {$\textbf{66.4}$}  \\
    \bottomrule
    \end{tabular}
    }
    
    \label{tab:main_tab_reb_mmd}
    \vspace{-12pt}
\end{table}

\section{Comparison with DG and additional experiments.}
Tab. \ref{tab:main_tab_reb_dg} includes additional results from two SOTA methods in DG Re-ID: TransMatcher \cite{liao2021transmatcher} and PAT \cite{ni2023part}. 
We compare these methods with Fed-Protoid (ViT) presented in Tab. 2 of the main paper. 
In all configurations, Fed-Protoid (ViT) outperforms the other methods.
These results are further improved using Fed-Protoid++, which incorporates the LUP large-scale dataset during pre-training instead of being initialized by ImageNet.
\begin{table}[h]
    \centering
    \vspace{-19pt}
    \caption{Comparison between Fed-Protoid and DG methods.}
    \resizebox{0.8\columnwidth}{!}{
    \begin{tabular}{lccccc}
    \toprule
    \multirow{2}{*}{\textbf{Method}} & \multirow{2}{*}{\textbf{Type}} & \multicolumn{2}{c}{\textbf{MS} $\to$ \textbf{M}} & \multicolumn{2}{c}{\textbf{MS} $\to$ \textbf{C}} \\ 
    & & \textbf{mAP} & \textbf{Rank-1}  & \textbf{mAP} & \textbf{Rank-1}  \\
    \midrule
    TransMatcher \cite{liao2021transmatcher} & DG & {$52.0$} & {$80.1$}  & {$22.5$} & {$23.7$}  \\
    PAT \cite{ni2023part} & DG & {$47.3$} & {$72.2$}  & {$25.1$} & {$24.2$}   \\ \cdashline{1-6}
    Fed-Protoid (ViT)~(\textcolor{red}{ours})& UDA & {${\underline{52.4}}$} & {${\underline{80.6}}$}& {${\underline{27.5}}$} & {${\underline{26.6}}$}  \\
    Fed-Protoid++~(\textcolor{red}{ours})& UDA & {$\textbf{61.7}$} & {$\textbf{82.6}$}& {$\textbf{43.8}$} & {$\textbf{42.4}$}  \\
    \bottomrule
    \end{tabular}
    }
    
    \label{tab:main_tab_reb_dg}
    \vspace{-8pt}
\end{table}

\end{document}